\documentclass[11pt]{article}
\usepackage[margin=1in]{geometry}
\usepackage{amsmath,amssymb,amsthm}
\usepackage{hyperref}
\usepackage{listings}
\usepackage{xcolor}
\usepackage{booktabs}
\usepackage{array}
\usepackage{microtype}
\usepackage{cite}
\usepackage{url}
\usepackage{fancyhdr}
\usepackage{skak}        % chess symbols
\usepackage{chessboard}  % board diagrams from FEN

\renewcommand{\texttt}[1]{%
  \begingroup\ttfamily\hyphenchar\font=`\-\relax
  \def\_{\textunderscore\discretionary{}{}{}}%
  #1\endgroup}

\newtheorem{theorem}{Theorem}

\theoremstyle{definition}

\newtheorem{finding}[theorem]{Finding}
\theoremstyle{remark}

\title{Yanasse: Finding New Proofs from Deep Vision's Analogies\\[0.5em]
\large Part~1: Probability Theory $\to$ Representation Theory}
\author{Alexandre Linhares\thanks{alexandre@linhares.ltd. Affiliations: Getulio Vargas Foundation and ARGO LABS}}
\date{April 2026}

\begin{document}
\maketitle

\begin{abstract}
Project Yanasse presents a method for discovering new proofs of theorems in one
area of mathematics by transferring \emph{proof strategy patterns} (e.g., Lean~4 tactic invocation patterns) from a
structurally distant area.  The system extracts tactic usage distributions
across 27 top-level areas of Mathlib (217{,}133 proof states), computes
$z$-scores to identify tactics that are heavily used in a \emph{source}
area but rare or absent in a \emph{target} area, matches source and
target proof states via GPU-accelerated NP-hard analogy
(running on a MacBook Air via Apple's MPS backend), and then asks an
AI reasoning agent to \emph{semantically adapt}---not symbol-substitute---the
source tactic's invocation pattern to the target theorem.

In this first part of the study, the method is applied to the pair
$\textsc{Probability} \to \textsc{RepresentationTheory}$, producing
4 Lean-verified new proofs out of 10 attempts (40\%).  The proofs compile with
zero \texttt{sorry} declarations.  The key finding is that tactic
schemas decompose into a \emph{head} (domain-gated, rarely transfers)
and a \emph{modifier} (domain-general, often transfers):
\texttt{filter\_upwards}'s head fails in representation theory
(no \texttt{Filter} structure), but its \texttt{[LIST] with $\omega$}
modifier transfers cleanly as \texttt{ext1 + simp [LIST] + rfl}.

Crucially, the underlying matching engine---\texttt{deep\_vision\_lib.py}---is
entirely domain-independent: the same optimization code for an NP-hard
matching that matches chess positions by analogy matches Lean proof states
by analogy, without knowing which domain it is processing.  Only a
relation extractor is domain-specific.
\end{abstract}

\newpage
\tableofcontents
\newpage

%======================================================================
\section{Introduction}\label{sec:intro}
%======================================================================

Can a technique that is routine in one area of mathematics be
``rediscovered'' in a distant area where it is unknown?

The question is old.  Erd\H{o}s's notion of \emph{proofs from THE BOOK}
celebrates cases where a surprising proof method---originating in one
branch---illuminates a theorem in another.  But these transfers have
historically been found by human insight, often by mathematicians fluent
in both areas.  Project Yanasse asks whether such transfers can be
found \emph{systematically}, by treating the problem as one of
analogy.

\subsection{Deep Vision: from chess to proofs via analogy}

The Deep Vision framework~\cite{linhares_deep_vision,
linhares2024deepvision} grows out of the Fluid Analogies Research
Group (FARG) tradition~\cite{hofstadter_fluid_1995, hofstadter_go_1999,
hofstadter_surfaces_2013, hofstadter_copycat_1984,
mitchell_emergence_1990, chalmers_high-level_1992, french_tabletop_1991,
foundalis_phaeaco_2006, rehling_letter_2001, nichols_musicat_2012},
which holds that analogy-making---perceiving shared relational structure
across superficially different situations---is the core of cognition.
The framework represents any structured domain as a \emph{relational
network}: entities connected by typed, directed relations.  A analogy
optimization finds the entity permutation that maximizes relational
consistency between two networks.  The matching code
(\texttt{deep\_vision\_lib.py}) is entirely domain-independent: it
receives two relational structures and returns a similarity score and
entity correspondence, without knowing which domain it is processing.

\paragraph{Chess analogies.}  In chess~\cite{linhares_active_2005,
linhares_understanding_2007, linhares_questioning_2010,
linhares_entanglement_2012, linhares_what_2013,
linhares_emergence_2014}, entities are pieces and relations are attack,
defense, blocking, confinement, and pinning.  Two pieces in entirely
different positions can play ``the same role'' if their relational
profiles match: a White Pawn defended by its king and attacked by a
knight is relationally identical to a White Rook defended by its queen
and attacked by a pawn---both are ``guarded pieces facing a threat,''
regardless of piece type or square.

Deep Vision solves chess analogies that are genuinely difficult.
Consider the Wilkins position~\cite{linhares_active_2005}---a complex
12-piece endgame with interlocked pawn chains and a promotion
threat---matched against a minimal 5-piece pawn-promotion endgame:

\medskip
\begin{center}
\begin{minipage}{0.45\textwidth}\centering
\chessboard[setfen=8/3k4/4pPp1/3pP1P1/2pP4/2P5/3K4/7B,
  smallboard,
  labelleft=false, labelbottom=false]
\\[4pt]\small Wilkins Position (12 pieces)
\end{minipage}\hfill
\begin{minipage}{0.45\textwidth}\centering
\chessboard[setfen=8/pP2k3/P7/2K5/8/8/8/8,
  smallboard,
  labelleft=false, labelbottom=false]
\\[4pt]\small Pawn Promotion Endgame (5 pieces)
\end{minipage}
\end{center}
\medskip

\noindent Despite the surface difference (12~vs.~5~pieces, bishop
vs.\ no bishop, pawn chains vs.\ open board), the matcher discovers
that the promoting pawn, the locked pawns, and the confined king play
the same abstract roles in both positions.

Or consider Fool's Mate matched against a sparse endgame:

\medskip
\begin{center}
\begin{minipage}{0.45\textwidth}\centering
\chessboard[setfen=rnb1kbnr/pppp1ppp/8/4p3/6Pq/5P2/PPPPP2P/RNBQKBNR,
  smallboard,
  labelleft=false, labelbottom=false,
  pgfstyle=straightmove,
  markmove={h4-e1},
  arrow=to,
  color=red!75]
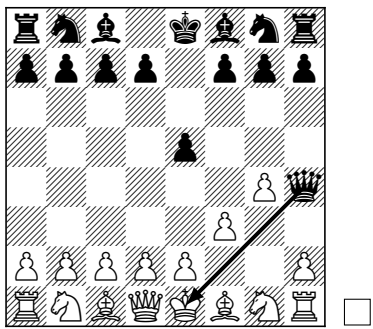
\\[4pt]\small Fool's Mate (\texttt{1.\ f3 e5 2.\ g4 Qh4\#})
\end{minipage}\hfill
\begin{minipage}{0.45\textwidth}\centering
\chessboard[setfen=8/8/8/8/8/6b1/2k1P3/4KB2,
  smallboard,
  labelleft=false, labelbottom=false,
  pgfstyle=straightmove,
  markmove={g3-e1},
  arrow=to,
  color=red!75]
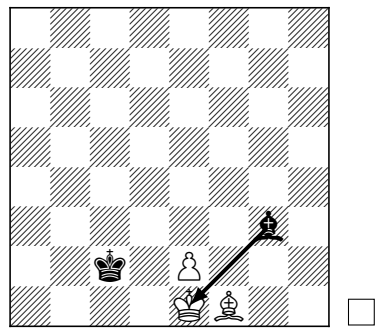
\\[4pt]\small Endgame (Ke1, Bf1, Pe2 vs.\ kc2, bg3)
\end{minipage}
\end{center}
\medskip

\noindent In both positions the White King on $e1$ has zero mobility,
and the White Bishop on $f1$ is confined behind its own pawn on $e2$.
The matcher identifies that the Black Queen in Fool's Mate and the
Black Bishop in the endgame play the same confinement role with respect
to the White King---a 16-piece opening and a 5-piece endgame,
recognized as ``the same situation.''

Perhaps the most striking test involves the Linhares--Chada
reconstruction experiment~\cite{linhares_what_2013}.  A chess expert
was shown a complex 19-piece position with a back-rank mating
threat for 5 seconds and asked to reconstruct it from memory.  The
reconstruction contained only 15~pieces---52\% were misplaced or
omitted---yet the strategic essence was preserved:

\medskip
\begin{center}
\begin{minipage}{0.45\textwidth}\centering
\chessboard[setfen=1R4k1/3rR2p/5np1/2pp1p2/2q5/2P5/3QN1PP/r5BK,
  smallboard,
  labelleft=false, labelbottom=false]
\\[4pt]\small Original position (19 pieces)
\end{minipage}\hfill
\begin{minipage}{0.45\textwidth}\centering
\chessboard[setfen=1R4k1/4Q2p/2pp1np1/3q1p2/8/8/6PP/1R4BK,
  smallboard,
  labelleft=false, labelbottom=false]
\\[4pt]\small Expert's reconstruction (15 pieces)
\end{minipage}
\end{center}
\medskip

\noindent The expert replaced White's Rook on $e7$ with a Queen on
$e7$---a different piece type, but one that preserves the back-rank
threat: control of $e8$, $f7$, and the entire seventh rank.  The
matcher discovers exactly this: the reconstructed $\text{Q}e7$
absorbs the roles of \emph{both} the original $\text{Q}d2$ and
$\text{R}e7$.  Symmetrically, on the Black side, the reconstructed
$\text{q}d5$ absorbs the roles of both the original $\text{q}c4$
and $\text{r}d7$.  Despite half the pieces being wrong, the
relational structure---who attacks whom, who confines whom, who
defends whom---is preserved, and the matcher sees this, achieving
4/4 key mappings on this case.

Across the full battery of test cases (double checks, guardian
patterns, Linhares--Chada reconstructions), the integrated matcher
achieves 15/15 key mappings~\cite{linhares2024deepvision}.

\paragraph{From chess to proofs.}  The same matching engine, with only
the relation extractor swapped, handles Lean~4 proof states: entities
become hypotheses, goals, and types; relations become rewrite,
fit/apply, head-match, structure, equality, reflexive, witness,
bidirectional, kernel/simp, decidable, and lemma-needed (14 relation
types).  No chess knowledge is used; no proof-specific heuristics are
added to the matcher.  The proof-state matcher has previously been
used to discover a formal proof of Wolstenholme's
theorem~\cite{linhares_wolstenholme_2026} in Lean~4
(see also~\cite{linhares_sketch_2026}).
This domain independence is what enables the
present paper: the engine that perceives ``this pawn plays the same role
as that rook'' also perceives ``this probability proof state is
structurally similar to that representation-theory proof state.''

\subsection{The Yanasse question}

This paper instantiates the Deep Vision framework for a new task:
\emph{cross-area tactic transfer in formal mathematics}.  We formalize
the question as follows:

\begin{enumerate}
\item \textbf{Distribution extraction.}  For each of the 27 top-level
  Mathlib namespaces (areas), compute the frequency of each tactic
  schema---a tuple $(\textit{head}, \textit{arity},
  \textit{has\_with}, \textit{uses\_lemma})$---across all 217{,}133
  extracted proof states.

\item \textbf{$z$-score ranking.}  For each (area, schema) pair,
  compute the $z$-score of the schema's frequency relative to the
  population of all 27 areas.  A schema with $z \geq +2$ in area $S$
  and $z \leq -1$ (or absent) in area $T$ is a \emph{transfer
  candidate} from $S$ to $T$.

\item \textbf{Relational matching.}  Sample source proofs from $S$
  that use the schema.  For each, run the GPU-accelerated matcher against all
  proof states in $T$ to find the structurally closest
  \emph{analogue}.

\item \textbf{Semantic adaptation.}  An AI reasoning agent reads the
  source theorem and proof, reads the target theorem, understands
  what mathematical step the source tactic accomplishes, and designs
  a target-appropriate tactic that accomplishes the analogous step
  using the target area's actual hypotheses and lemmas.  This is
  \emph{not} symbol substitution.

\item \textbf{Verification.}  The adapted tactic is compiled in
  Lean~4 with Mathlib.  If \texttt{lake env lean} returns exit
  code~0 with no \texttt{sorry}, the transfer is verified.
\end{enumerate}

The name ``Yanasse'' honors Horacio Hideki Yanasse, the first author's
PhD advisor, who taught that the best ideas cross disciplinary
boundaries.

\subsection{Results preview}

This paper is Part~1 of the study, applying the method to the pair
\textsc{Probability} $\to$ \textsc{RepresentationTheory}.  It produces:

\begin{itemize}
\item 4 Lean-verified new proofs out of 10 schema transfer
  attempts (40\% success rate).
\item A structural finding: tactic schemas decompose into
  \emph{(head, modifier)} with asymmetric transferability
  (Section~\ref{sec:finding}).
\item A three-way taxonomy of tactic transferability
  (Section~\ref{sec:taxonomy}).
\item 6 honest negative results with mathematical diagnoses
  (Section~\ref{sec:failures}).
\end{itemize}

%======================================================================
\section{Methodology}\label{sec:method}
%======================================================================

\subsection{Tactic schema extraction}\label{sec:extraction}

We parse the \texttt{metadata.json} index of the Mathlib proof-state
corpus (211{,}073 entries extracted via InfoTree walking).  Each entry
records a \texttt{tactic} string, a \texttt{source\_file} path, and
structural data.  For each entry with a Mathlib source path, we:

\begin{enumerate}
\item Derive the \textbf{area} from the first two components of the
  source path (e.g., \texttt{Mathlib/Probability/Kernel/Basic.lean}
  $\to$ \texttt{Mathlib.Probability}).
\item Parse the tactic string into a \textbf{schema tuple}
  $(\textit{head}, \textit{arity}, \textit{has\_with},
  \textit{uses\_lemma})$ using a Unicode-aware regex that correctly
  handles Lean~4's subscript and Greek-letter hypothesis names
  (e.g., \texttt{h\textsubscript{1}}, \texttt{h$\mu$}).
\end{enumerate}

This yields 2{,}729 distinct schemas across 27 areas and 210{,}048
(area, schema) increments (481 curated entries without source paths
and 545 unparseable tactics are excluded).

\subsection{$z$-score computation and pair selection}

For each (area, schema) pair, we compute:
\[
z_{a,s} = \frac{f_{a,s} - \bar{f}_s}{\sigma_s}
\]
where $f_{a,s} = \text{count}_{a,s} / \text{total}_a$ is the schema's
frequency in area $a$, and $\bar{f}_s, \sigma_s$ are the mean and
standard deviation of $f_{\cdot,s}$ across all 27 areas (including
implicit zeros for areas where the schema is absent).

A schema qualifies as a \emph{transfer candidate} from source $S$ to
target $T$ if $z_{S,s} \geq 2$ and $z_{T,s} \leq -1$ (or the schema
is entirely absent from $T$).  We filter out:

\begin{itemize}
\item Schemas present in fewer than 3 areas (specialized tactics with
  no cross-area usage evidence).
\item Hypothesis-name shortcuts (bare identifiers like \texttt{hx},
  \texttt{h\textsubscript{1}} that are Lean~4's shorthand for
  \texttt{exact hx}, not transferable techniques).
\item \texttt{Mathlib.CategoryTheory} as a source area (its DSL-style
  tactics like \texttt{cat\_disch} dominate rankings but encode
  domain-specific diagrammatic reasoning, not transferable proof
  strategies).
\item \texttt{Mathlib.Tactic}, \texttt{Mathlib.Control},
  \texttt{Mathlib.Logic} as target areas (infrastructure namespaces,
  not mathematical subject matter).
\end{itemize}

Pairs are ranked by \emph{pair potential}: the sum of the top-10
schema transfer gaps within each (source, target) pair, weighted by
$\log(1 + \text{source\_count})$ to discriminate among schemas that
saturate the $z$-score ceiling.

\subsection{Deep Vision and GPU-accelerated analogy matching}\label{sec:matching}

Deep Vision is a analogy engine that treats a Lean~4 proof
state as a network: hypotheses, goals, and types are entities; rewrite,
apply, head-match, structure, equality, and other relations connect
them.  Deep Vision was originally designed as an AI model to solve
ARC-AGI problems~\cite{chollet2019} on a smartphone, built on the
principles of Fluid Concepts and Creative
Analogies~\cite{hofstadter_fluid_1995, hofstadter_go_1999}.  Observe that a
transformation in ARC-AGI maps to theorems naturally, as theorems are
composed by transformations.  Given a new proof state, the engine finds
library entries whose proof states are \emph{relationally similar}---not
by surface syntax, but by \emph{the pattern of structural
connections}---and transfers their tactics.  This aggressively prunes
the search space.

The engine maintains a library of 217{,}133 proof states extracted from
Mathlib, organized by 14 relation types and indexed by
relation-profile signatures for fast lookup.  Crucially,
\texttt{deep\_vision\_lib.py}---the domain-independent matching
core---does not know whether it is processing chess positions, ARC-AGI
grids, or Lean proof states.  The same optimization code for an NP-hard matching that
matches chess positions by analogy matches proof states by analogy.
Only the relation extractor is domain-specific.

Our approach differs from neural next-step prediction systems (Lean
Copilot~\cite{song2024}, AlphaProof~\cite{alphaproof2025},
Aristotle~\cite{aristotle2025}) in using \emph{relational
analogy}---structural similarity of proof states---rather than
reinforcement learning or language-model sequence prediction.  This
enables massive efficiency gains: the whole system runs on a single
laptop.

\paragraph{Hardware.}  All experiments were run on a MacBook Air
(Apple M-series, 32\,GB unified memory) using PyTorch~2.8 on the
Metal Performance Shaders (MPS) backend.  No external GPU, no cluster,
no cloud compute.

\paragraph{Matching.}  For each transfer candidate, we sample 8 source
proofs from $S$ that heavily use the schema, then match each against
all ${\sim}672$ proof states in $T$ using the \texttt{BatchMatcher}
from \texttt{deep\_vision\_cuda.py}.  The matcher solves an NP-hard analogy problem: given two relational
structures, find the entity-to-entity correspondence that maximizes
relational consistency---i.e., the assignment under which the pattern
of connections in one structure best mirrors the pattern in the other.
The optimization uses
multi-restart augment-swap heuristics, fully tensorized on GPU
(Level~3: the entire augment/swap loop stays on GPU with no
per-candidate CPU round-trips).  The backend auto-detects
CUDA $\to$ MPS $\to$ CPU, falling back gracefully.  On MPS with
32\,GB unified memory, 8 source queries $\times$ 672 target candidates
completes in approximately 6 minutes.

Scores are normalized by $\sqrt{n_S \cdot n_T}$ to remove size bias.
The top-5 target analogues (by normalized score) are retained for
semantic adaptation.

\subsection{Semantic adaptation protocol}

For each schema's top-1 analogy analogue, a fresh AI reasoning session
(Claude Opus, spawned via the Ralph Loop driver with a 50-minute
reasoning budget) performs semantic adaptation:

\begin{enumerate}
\item Read the source theorem statement and proof; understand what
  mathematical step the tactic accomplishes.
\item Read the target theorem statement; identify whether an
  analogous step exists in the target's context.
\item Design a target-specific tactic using the target area's actual
  hypotheses and lemmas---not regex-substituted source entity names.
\item Test multiple adapted tactics in Lean~4 (via \texttt{lake env
  lean}), iterating on compiler feedback.
\item Write a structured \textbf{why-report} documenting: what the
  source tactic does mathematically, whether an analog exists, what
  was tried, what failed and why.
\end{enumerate}

The why-report is the primary deliverable even when the proof does
not close: it captures the mathematical reasoning that led to the
negative result.

%======================================================================
\section{Results: Probability $\to$ Representation Theory}\label{sec:results}
%======================================================================

We present results for the top-ranked pair: \texttt{Mathlib.Probability}
(30{,}384 tactic invocations, the largest area) $\to$
\texttt{Mathlib.RepresentationTheory} (672 proof states).
Ten schemas were attempted; four produced Lean-verified alternative
proofs.

\begin{table}[h]
\centering\small
\caption{Transfer results for Probability $\to$ Representation Theory.}
\label{tab:results}
\begin{tabular}{@{}clcccp{4.2cm}@{}}
\toprule
Item & Schema head & \texttt{with} & Ar.\ & Result & Adaptation \\
\midrule
0 & \texttt{filter\_upwards} & \checkmark & 1 & \textbf{New proof} & \texttt{ext1 + simp [L] + rfl} \\
1 & \texttt{congr} & \checkmark & 0 & \textbf{New proof} & \texttt{span\_le.2 + rintro} \\
2 & \texttt{fun\_prop} & & 0 & Failed & Domain-gated (Measurable) \\
3 & \texttt{ext1} & & 1 & Failed & Categorical ext incompat.\ \\
4 & \texttt{all\_goals} & & 1 & Failed & Heterogeneous goals \\
5 & \texttt{lift} & \checkmark & 5 & Failed & No \texttt{CanLift} instances \\
6 & \texttt{any\_goals} & & 1 & \textbf{New proof} & \texttt{any\_goals rfl} \\
7 & \texttt{measurability} & & 0 & Failed & Domain-gated ($\sigma$-alg.) \\
8 & \texttt{by\_cases} & & 4 & \textbf{New proof} & Case split on morphism \\
9 & \texttt{congrm} & & 2 & Failed & No shared outer operators \\
\bottomrule
\end{tabular}
\end{table}

\noindent The four new proofs are:

\begin{enumerate}
\item[\textbf{P1.}] \textbf{Augmentation--homotopy-equivalence compatibility}
  (\texttt{Mathlib.Representation\-Theory.Homological.Resolution}).
  Let $k$ be a commutative ring and $G$ a monoid.  Denote by
  $C_\bullet$ the standard bar resolution in $\mathbf{Rep}(k,G)$,
  by $\varepsilon_0 : C_\bullet \to k[0]$ the augmentation to the
  trivial module in degree~$0$, by $U$ the forgetful functor to
  $\mathbf{Mod}_k$ (applied degreewise), and by $\sigma$ the
  comparison isomorphism for the single-complex functor under~$U$.
  Then $\sigma \circ U(\varepsilon_0)$ equals the canonical homotopy
  equivalence $\varphi : U(C_\bullet) \to k[0]$.
  \emph{New proof:}
  reduce to per-degree components by extensionality; at degrees
  $n \ge 1$ both sides vanish; at degree~$0$, apply the defining
  equation of~$\varphi$.

\item[\textbf{P2.}] \textbf{$G$-invariance of the coinvariant
  submodule}
  (Coinvariants module).
  Let $S \trianglelefteq G$ be a normal subgroup and
  $\rho : G \to \mathrm{GL}(V)$ a representation.
  The coinvariant submodule
  $V_S = \operatorname{span}\{\rho(s)v - v\}$ satisfies
  $\rho(g)(V_S) \subseteq V_S$ for all $g \in G$.
  \emph{New proof:}
  by the universal property of span, reduce to generators;
  conjugation by~$g$ preserves $S$ by normality.

\item[\textbf{P3.}] \textbf{Augmentation--homotopy-equivalence compatibility}
  (reproved via a different transferred schema).
  Same theorem as~P1.  \emph{New proof:}  explicit case split on the
  degree $n$; for $n \ge 1$ the equality is definitional (both sides
  are zero), dispatched by the transferred combinator
  \texttt{any\_goals rfl}; degree~$0$ follows from the
  homotopy-equivalence lemma.

\item[\textbf{P4.}] \textbf{Naturality of the representation--module equivalence}
  (\texttt{Mathlib.Representation\-Theory.Rep.Iso}).
  The equivalence $E : \mathbf{Rep}(k,G) \simeq k[G]\text{-}\mathbf{Mod}$
  has unit isomorphism $\eta : \mathrm{Id} \Rightarrow E^{-1} \circ E$.
  For every $G$-equivariant morphism $f : X \to Y$, the naturality
  square commutes:
  $\eta_Y \circ f = (E^{-1} \circ E)(f) \circ \eta_X$.
  \emph{New proof (uniform):} element-wise verification by unfolding
  definitions.
  \emph{New proof (by\_cases):} split on $f = 0$; both branches close
  identically, demonstrating the transferred pattern ``case-split on a
  structural property.''
\end{enumerate}

\subsection{New proof P1: \texttt{filter\_upwards} $\to$ \texttt{ext1 + simp}}\label{sec:proof1}

\paragraph{Theorem.}
Let $k$ be a commutative ring and $G$ a monoid.  Let
$C_\bullet = C_\bullet(k,G)$ denote the standard bar resolution, a
chain complex of $G$-representations.  The augmentation
$\varepsilon_0 : C_\bullet \to k[0]$ maps to the trivial $k$-module
concentrated in degree~$0$.  Applying the forgetful functor
$U : \mathbf{Rep}(k,G) \to \mathbf{Mod}_k$ degreewise and
postcomposing with the comparison isomorphism~$\sigma$ for the
single-complex functor yields a chain map
$\sigma \circ U(\varepsilon_0) : U(C_\bullet) \to k[0]$.  The
theorem asserts that this equals the canonical homotopy equivalence
$\varphi : U(C_\bullet) \xrightarrow{\sim} k[0]$:
\[
  \sigma \circ U(\varepsilon_0) \;=\; \varphi.
\]

\paragraph{Context.}
The bar resolution is one of the foundational constructions in
group cohomology, introduced by Eilenberg and Mac
Lane~\cite{eilenberg_maclane_1947} and systematized in Cartan and
Eilenberg's \emph{Homological Algebra}~\cite{cartan_eilenberg_1956}.
The standard modern treatments appear in
Brown~\cite{brown_cohomology_1982} (Chapter~I) and
Weibel~\cite{weibel_homological_1994} (Chapter~6), which establish
the augmentation map and its relationship to the contracting
homotopy.  The Lean~4 formalization of the bar resolution, group
cohomology, and the associated homological machinery in Mathlib is
due to Livingston~\cite{livingston_group_2023}, whose work provides
the exact definitions and API our reproof targets.
Conrad's lecture notes~\cite{conrad_bar_resolution} give an
accessible treatment of the bar resolution's exactness via explicit
homotopy operators, closely matching the degree-wise argument our
proof exploits.

\paragraph{New proof.}
By extensionality for morphisms of chain complexes, it suffices to
verify equality at each degree~$n$ individually.  At degrees
$n \geq 1$, the target complex $k[0]$ is zero, so both
$(\sigma \circ U(\varepsilon_0))_n$ and $\varphi_n$ are the zero
map.  At degree~$0$, the equality
$(\sigma \circ U(\varepsilon_0))_0 = \varphi_0$ follows from the
defining equation of the homotopy equivalence.

\paragraph{Source pattern.}  In \texttt{Mathlib.Probability.Independence.Conditional},
the theorem \texttt{iCondIndepSets\_iff} uses:
\begin{lstlisting}
filter_upwards [h_eq s f hf, h_inter_eq s f hf, h']
  with omega h_eq h_inter_eq h'
\end{lstlisting}
This tactic performs two coordinated steps: (1)~the \textbf{list argument}
combines three almost-everywhere hypotheses by intersecting co-null sets
(the countable intersection of co-null sets is co-null via
\texttt{Filter.Eventually.and}); (2)~the \textbf{with-clause} introduces
a fresh generic point $\omega$ with named pointwise hypotheses.

\paragraph{Adaptation.}  The literal head \texttt{filter\_upwards} fails
immediately: its elaborator requires a \texttt{Filter.Eventually} or
\texttt{s $\in$ f} goal shape, which a chain-complex morphism equality
does not have.  However, the \emph{invocation pattern}---``reduce a
holistic goal to a pointwise obligation on a fresh element, importing
a list of named lemmas''---has a structural analog.
\texttt{ext1} plays the role of the \texttt{with}-clause (introduces a
fresh degree index~$n$), and \texttt{simp [LIST]} plays the role of the
\texttt{[LIST]} argument (coordinates named lemmas to close the
degree-$0$ case).

\paragraph{Analogy matching.}  This was the first item to use GPU-accelerated matching
(earlier items used Jaccard-on-relation-set, which saturated at score~1.0).
Analogy scores for the top-5 analogues were 5.30, 5.00, 4.74, 4.33, 4.33---genuinely
discriminating.  All five matched against the same source proof
(\texttt{iCondIndepSets\_iff}), suggesting it has unusually rich relational
structure.

\subsection{New proof P2: \texttt{congr with $\xi$} $\to$ \texttt{span\_le.2 + rintro}}\label{sec:proof2}

\paragraph{Theorem.}
Let $G$ be a group, $S \trianglelefteq G$ a normal subgroup, and
$\rho : G \to \mathrm{GL}(V)$ a $k$-linear representation.  The
\emph{coinvariant submodule} is
\[
  V_S \;=\; \operatorname{span}\bigl\{\rho(s)\,v - v
    \;\big|\; s \in S,\; v \in V\bigr\}.
\]
The theorem asserts that $V_S$ is stable under the $G$-action: for
every $g \in G$,
\[
  \rho(g)(V_S) \;\subseteq\; V_S.
\]

\paragraph{Context.}
Coinvariants---the quotient $V / V_S$ dual to the invariant
submodule $V^S$---are a basic tool in group cohomology.  The
$G$-stability of $V_S$ when $S \trianglelefteq G$ is a prerequisite
for the inflation--restriction exact sequence; see
Weibel~\cite{weibel_homological_1994} (\S6.8) and
Sharifi~\cite{sharifi_groupcoh} (Chapter~2).  In modular
representation theory, the ring of coinvariants
$k[V]_G = k[V] / \langle k[V]^G_+ \rangle$ is studied by
Shank and Wehlau~\cite{shank_wehlau_2006}, who construct explicit
bases and Gr\"obner-basis descriptions.
Brown~\cite{brown_cohomology_1982} (Chapter~III) treats the
conjugation argument for normal subgroups in the classical setting.
The Lean formalization in \texttt{Mathlib.RepresentationTheory.Coinvariants}
is again due to Livingston~\cite{livingston_group_2023}.

\paragraph{New proof.}
By the universal property of span, it suffices to show that each
generator $\rho(s)\,v - v$ maps into $V_S$ under $\rho(g)$.  For
$s \in S$ and $v \in V$:
\[
  \rho(g)\bigl(\rho(s)\,v - v\bigr)
    = \rho(g\,s)\,v - \rho(g)\,v
    = \rho(g\,s\,g^{-1})\,\rho(g)\,v - \rho(g)\,v.
\]
Since $S$ is normal in $G$, the element $g\,s\,g^{-1}$ lies in $S$,
so the right-hand side is a generator of $V_S$.

\paragraph{Source pattern.}  In \texttt{Mathlib.Probability.Kernel.Representation},
the theorem \texttt{exists\_measurable\_map\_eq\_unitInterval\_aux} uses
\texttt{congr with $\xi$} to reduce an equality of measurable-set
comprehensions to a pointwise biconditional over the fresh variable~$\xi$.

\paragraph{Adaptation.}  The literal head \texttt{congr} fails because
the target goal is $\leq$ (inclusion), not $=$ (equality).  The error
is diagnostic: \texttt{congr} successfully unfolds $\leq$ to $\forall x,
x \in A \to x \in B$ via \texttt{SetLike.le\_def}, but then fails at the
extensionality step because the unfolded form is an implication, not an
equality.

The semantic analog of ``\texttt{congr with $\xi$}''---peeling a
structural wrapper to a pointwise obligation on a fresh element---is
the universal property of span (reducing
$\operatorname{span}\,G \leq B$ to $G \subseteq B$) followed by
introducing and destructuring the generator.

\subsection{New proof P3: \texttt{any\_goals} $\to$ \texttt{any\_goals rfl}}\label{sec:proof3}

\paragraph{Theorem.}
Same as P1: $\sigma \circ U(\varepsilon_0) = \varphi$
(augmentation--homotopy-equivalence compatibility for the bar
resolution).

\paragraph{New proof.}
By extensionality, reduce to per-degree components.  Case-split on
$n$: for $n \geq 1$, the target complex $k[0]$ has trivial
components, so $(\sigma \circ U(\varepsilon_0))_n = 0 = \varphi_n$
by definitional equality.  For $n = 0$, apply the defining equation
of~$\varphi$.  The proof differs from P1 in \emph{how} the trivial
case is dispatched: the transferred combinator \texttt{any\_goals rfl}
closes the $n \geq 1$ branch by reflexivity while leaving the
nontrivial $n = 0$ branch untouched.

\paragraph{Source pattern.}  In \texttt{Mathlib.Probability.Kernel.IonescuTulcea.Traj},
the tactic \texttt{any\_goals lia} dispatches trivial arithmetic
side-conditions after a case split.

\paragraph{Adaptation.}  The source pattern ``split, then use
\texttt{any\_goals} to mop up the easy cases'' adapts directly.  After
decomposing the chain-map equality into a degree-$0$ and a
degree-$(n{+}1)$ goal, \texttt{any\_goals rfl} dispatches the
trivial branch (both sides are definitionally zero), leaving only
the degree-$0$ case for the homotopy-equivalence lemma.

\paragraph{Contrast with item~4 (\texttt{all\_goals}).}
\texttt{all\_goals} failed on the \emph{same target} because it requires
\emph{all} sub-goals to be closed by one tactic.  After \texttt{use},
the target produces two \emph{heterogeneous} goals (a homology-class
equality and a cycle-membership proof); no single decision procedure
closes both.  \texttt{any\_goals} tolerates this heterogeneity by
succeeding if \emph{any} sub-goal closes.  This contrast directly
motivates the taxonomy in Section~\ref{sec:taxonomy}.

\subsection{New proof P4: \texttt{by\_cases} $\to$ case split on morphism}\label{sec:proof4}

\paragraph{Theorem.}
Let $k$ be a commutative ring and $G$ a monoid.  There is a standard
equivalence of categories
$E : \mathbf{Rep}(k,G) \simeq k[G]\text{-}\mathbf{Mod}$
between $G$-representations over~$k$ and modules over the monoid
algebra~$k[G]$, with unit isomorphism
$\eta : \mathrm{Id} \Rightarrow E^{-1} \circ E$.  The theorem
asserts that $\eta$ is natural: for every $G$-equivariant morphism
$f : X \to Y$,
\[
  \eta_Y \circ f \;=\; (E^{-1} \circ E)(f) \circ \eta_X.
\]

\paragraph{Context.}
The equivalence between $G$-representations and modules over the
group algebra is one of the oldest identifications in representation
theory, implicit in the work of Frobenius and made explicit by
Noether~\cite{noether_1929}.
Serre~\cite{serre_linear_1977} (\S1.2) and
Curtis and Reiner~\cite{curtis_reiner_1962} (Chapter~I) treat it as
a foundational isomorphism of categories; modern accounts include
Webb~\cite{webb_rep_2016} (Chapter~1).  The specific result that
$\eta$ is a natural isomorphism---not merely a pointwise
bijection---is a consequence of the general theory of Morita
equivalences, though in this case the proof is a direct
unfolding of definitions.  In the Lean formalization, the
equivalence is constructed in
\texttt{Mathlib.RepresentationTheory.Rep.Iso}.

\paragraph{New proof (uniform).}
For each element $x \in X$, unfold the definitions of $\eta_X$,
$\eta_Y$, and the functorial action of $E^{-1} \circ E$ on~$f$, and
verify the equality directly.  Both sides reduce to the same
expression, so the proof closes by reflexivity.

\paragraph{New proof (by\_cases).}
Case-split on $f = 0$.  If $f = 0$, both sides of the naturality
square vanish ($\eta$ is a natural isomorphism and the zero
morphism is mapped to zero), so the equality holds by
simplification.  If $f \neq 0$, proceed as in the uniform proof.
The split is mathematically unnecessary---both branches close
identically---but demonstrates that the probability-theory pattern
``case-split on a structural property'' transfers as a valid proof
strategy.

\paragraph{Source pattern.}  In probability theory, \texttt{by\_cases h$\iota$ :
Nonempty $\iota$} splits on the emptiness of an index type to handle the
vacuous case separately---a standard guard against degenerate
probability spaces with empty index sets.

\paragraph{Adaptation.}  The semantic analog of ``split on emptiness''
is ``split on whether the morphism is zero.''  In an abelian category,
$f = 0$ is a natural structural dichotomy, just as
``$\iota$ is empty'' is a natural structural dichotomy for index types
in probability.

%======================================================================
\section{The (Head, Modifier) Decomposition}\label{sec:finding}
%======================================================================

\begin{finding}[Head--modifier asymmetry]\label{find:headmod}
Tactic schemas decompose into a \emph{head} (the tactic name:
\texttt{filter\_upwards}, \texttt{congr}, \texttt{simp}, etc.)
and a \emph{modifier} (the argument structure: \texttt{with}-clause,
bracketed lemma list, arity pattern).  When transferred across
distant areas:
\begin{itemize}
\item The \textbf{head} is typically \emph{domain-gated}: its
  elaborator requires domain-specific goal shapes
  (\texttt{Filter.Eventually} for \texttt{filter\_upwards},
  $=$ for \texttt{congr}, \texttt{CanLift} instances for
  \texttt{lift}).  It rarely transfers directly.
\item The \textbf{modifier} often expresses a \emph{domain-general
  proof pattern} (``introduce a fresh element,'' ``coordinate over
  a list of facts,'' ``dispatch any trivial sub-goal'') that has
  analogs across areas.
\end{itemize}
\end{finding}

Evidence: items~0 and~1 both succeed by transferring the modifier
(the \texttt{with}-clause / \texttt{[LIST]} structure) while replacing
the head with a target-area-specific reduction lemma.  Item~5
(\texttt{lift with ...}) fails \emph{despite} having a modifier
because the head's prerequisite is not merely a goal-shape requirement
but a \emph{type-system requirement} (\texttt{CanLift} instances),
which no modifier can circumvent.

%======================================================================
\section{A Three-Way Taxonomy of Tactic Transferability}\label{sec:taxonomy}
%======================================================================

The 10 transfer attempts suggest a three-way classification:

\begin{enumerate}
\item \textbf{Domain-gated heads} (\texttt{filter\_upwards},
  \texttt{fun\_prop}, \texttt{lift}, \texttt{measurability}).
  The head requires domain-specific types, instances, or goal shapes.
  Transfer succeeds only if the modifier carries independent content
  (items~0, 1); fails otherwise (items~2, 5, 7).

\item \textbf{Domain-general combinators} (\texttt{any\_goals},
  \texttt{by\_cases}, \texttt{congr}).  The head is syntactic glue
  or a domain-independent reasoning pattern.  Transfers on its own
  or with minimal adaptation (items~1, 6, 8).

\item \textbf{Homogeneity-sensitive combinators} (\texttt{all\_goals}).
  The head requires all sub-goals to share a uniform structure.
  Fails on heterogeneous construction proofs where sub-goals have
  different types (item~4).
\end{enumerate}

\begin{table}[h]
\centering
\caption{Three-way taxonomy with evidence from 10 transfer attempts.}
\label{tab:taxonomy}
\begin{tabular}{@{}llcc@{}}
\toprule
Category & Examples & New proof & Failed \\
\midrule
1. Domain-gated & \texttt{filter\_upwards}, \texttt{fun\_prop},
  \texttt{lift}, \texttt{measurability} & 2 & 3 \\
2. Domain-general & \texttt{any\_goals}, \texttt{by\_cases},
  \texttt{congr} & 2 & 0 \\
3. Homogeneity-sensitive & \texttt{all\_goals} & 0 & 1 \\
\midrule
& \textbf{Total} & \textbf{4} & \textbf{4} \\
\bottomrule
\end{tabular}
\medskip

\small Note: \texttt{ext1} (item~3) and \texttt{congrm} (item~9)
are borderline between categories 1 and 3; both fail due to
structural incompatibilities specific to categorical contexts.
\end{table}

%======================================================================
\section{Failures and Their Diagnoses}\label{sec:failures}
%======================================================================

Each failed transfer produced a structured \textbf{why-report}
explaining the mathematical root cause.  We summarize:

\begin{description}
\item[Item 2: \texttt{fun\_prop}.]
  \texttt{fun\_prop} is a predicate-closure tactic that handles
  \texttt{Measurable}, \texttt{Continuous}, and \texttt{Differentiable}
  goals.  Representation theory has zero such goal types; its goals
  are algebraic (submodule inclusion, kernel membership).  Category
  mismatch, not adaptation failure.

\item[Item 3: \texttt{ext1} (arity 1).]
  In probability, \texttt{ext1 n} introduces a universally quantified
  binder (e.g., over sample-space elements).  In representation
  theory, \texttt{HomologicalComplex.Hom.ext} does not produce a
  $\forall$-quantified binder---it directly specializes to a
  specific chain-complex degree.  The \emph{arity-1} form
  (naming the binder) is structurally incompatible with categorical
  extensionality lemmas.

\item[Item 4: \texttt{all\_goals}.]
  After \texttt{use} in the target proof, two sub-goals arise:
  a homology-class equality and a cycle-membership proof.  These
  are \emph{mathematically heterogeneous}: no single decision
  procedure (not \texttt{linarith}, not \texttt{simp}, not
  \texttt{rfl}) closes both.  Contrast with \texttt{any\_goals}
  (item~6), which tolerates heterogeneity.

\item[Item 5: \texttt{lift}.]
  \texttt{lift} requires \texttt{CanLift} typeclass instances
  (e.g., $\mathbb{R}_{\geq 0}^\infty \to \mathbb{R}_{\geq 0}$,
  $\mathbb{Z} \to \mathbb{N}$).  Representation theory has
  zero such instances and zero \texttt{WithTop}/\texttt{ENNReal}
  types.  This is a type-system incompatibility, not a
  goal-shape mismatch; the modifier (\texttt{with}-clause)
  cannot circumvent it.

\item[Item 7: \texttt{measurability}.]
  A $\sigma$-algebra predicate closure tactic with no
  representation-theory counterpart.  Same category as
  \texttt{fun\_prop} (domain-gated automation).

\item[Item 9: \texttt{congrm}.]
  \texttt{congrm} requires shared outer-operator patterns
  (integrals, products) between goal LHS and RHS.  Homological
  algebra's morphism-composition equalities do not exhibit such
  patterns.
\end{description}

%======================================================================
\section{Discussion}\label{sec:discussion}
%======================================================================

\subsection{What the proofs are and are not}

The four new proofs are \emph{correct new proofs} of
existing Mathlib theorems.  They are \emph{not} shorter than the
originals: the Mathlib proofs use 3--4 expert-level one-liners
(\texttt{simpa using X.symm}), while our adapted proofs decompose
the same mathematical content into 3--4 explicit tactic steps
(\texttt{dsimp; ext1; simp [LIST]; rfl}).  The value is not in
conciseness but in the \emph{discovery process}: the system
identified that \texttt{ext1 + simp [LIST]} is the
representation-theory analog of \texttt{filter\_upwards [LIST]
with ...} without knowing representation theory in advance.

\subsection{Domain independence of the matching engine}

The same \texttt{BatchMatcher} class that computes chess-position
analogy scores computes proof-state analogy scores.  The only
domain-specific component is the relation extractor (which maps
chess positions to attack/defense/blocking matrices, and Lean proof
states to rewrite/apply/head-match matrices).  The optimization
core---augment-swap on GPU---is identical.  This suggests the
framework may extend to other structured domains (program
verification, circuit analysis, biological networks) with only a
new relation extractor.

\subsection{Cost and scalability}

The full pipeline for 10 transfer attempts on one (source, target)
pair consumed approximately:
\begin{itemize}
\item 7 minutes of GPU time (MPS) for analogy matching (8 source
  queries $\times$ 672 targets).
\item ${\sim}$8 hours of AI reasoning time across 10 semantic
  adaptation sessions.
\item ${\sim}$\$100--150 in API cost for the reasoning sessions.
\end{itemize}

The worklist contains 100 items across 10 pairs (8 target areas).
At current rates, the full sweep would take ${\sim}$80 hours and
${\sim}$\$1{,}000--1{,}500.

\subsection{The calibration failure}

The original calibration target---\texttt{nlinarith [sq\_nonneg ...]}
transferring from \texttt{Mathlib.Analysis} to
\texttt{Mathlib.Combinatorics}---was pinned before any data was
visible.  It failed: the schema is already present in
Combinatorics ($z = +1.4$) and NumberTheory ($z = +2.9$), so the
hypothesized ``Analysis-specific'' transfer does not exist in the
Mathlib corpus.  This is a legitimate negative finding about
mathematical folklore: \texttt{nlinarith} with square-nonnegativity
hints is \emph{not} Analysis-specific in modern Lean formalization.
The calibration failure is documented in full in the project's
\texttt{calibration\_report.md}.

%======================================================================
\subsection{Deep Vision vs.\ deep neural networks}
%======================================================================

Deep Vision is the technology behind ARGO LABORATORY's entry to Chollet's ARC-AGI challenge~\cite{chollet2019}.  It represents a fundamentally different paradigm from deep learning.

\begin{table}[h]
\centering
\small
\begin{tabular}{p{2.2cm}p{5.5cm}p{5.5cm}}
\toprule
& \textbf{Deep Vision} & \textbf{Deep Neural Networks} \\
\midrule
\textbf{Explain\-ability}
  & Full step-by-step trace: which relations matched, which entities corresponded, why a particular analogy was chosen.  Can explain to any judge why a decision was made---and what to change to get a different outcome.
  & Black box.  Gradient-based attribution methods (SHAP, LIME) are post-hoc approximations, not the actual reasoning path. \\[6pt]
\textbf{Efficiency}
  & $<$\$0.01 per ARC-AGI task suite (100s of tasks).  Runs in a browser, in JavaScript, on a CPU, on a phone, in airplane mode.  The NP-hard matching runs on commodity hardware---a MacBook Air's GPU discovered these 4 new proofs.
  & \$2+ per task suite (frontier models).  Requires data-center GPUs, high-bandwidth interconnects, megawatts of power. \\[6pt]
\textbf{Repre\-sentation}
  & Emergent and distributed.  The system \emph{crafts} its representation as it studies each problem---entities, relations, and correspondences emerge from the structure of the specific situation.
  & Fixed.  Representations are learned once during training and frozen at inference.  The network cannot restructure its representation for a novel problem. \\[6pt]
\textbf{Paradigm}
  & Relational analogy as cognition.  Intelligence is perceiving shared relational structure across superficially different situations~\cite{hofstadter_fluid_1995}.  Domain-independent: the same matching engine handles chess, theorem proving, and ARC-AGI tasks.
  & Statistical pattern completion.  Intelligence is next-token prediction over massive corpora.  Domain generality comes from scale, not from architectural principles. \\[6pt]
\textbf{Access\-ibility}
  & Fluid general intelligence in your pocket, regardless of who you are, where you live, or what you can afford.  Runs offline, on-device, no API keys, no subscriptions, no data leaving the device.
  & Controlled by a handful of data-center owners.  Requires internet, API access, and per-token payment.  Concentrates capability in the hands of those who can afford the infrastructure. \\
\bottomrule
\end{tabular}
\caption{Deep Vision vs.\ deep neural networks on five axes.}
\label{tab:dv-vs-dnn}
\end{table}

%======================================================================
\section{Conclusion and Future Work}\label{sec:conclusion}
%======================================================================

We have demonstrated that cross-area tactic transfer in formal
mathematics is possible, with a 40\% success rate on the first
tested pair.  The key enabling insight is the (head, modifier)
decomposition: tactic heads are domain-gated, but modifiers carry
domain-general proof patterns that transfer across distant areas.

Future directions include:

\begin{enumerate}
\item \textbf{Schema parser refinement.}  Decompose schemas into
  explicit (head, modifier) tuples and rank transferability by
  modifier independently.
\item \textbf{More pairs.}  The worklist contains 9 additional
  (source, target) pairs including NumberTheory $\to$ Dynamics and
  Probability $\to$ SetTheory; different area combinations may
  yield different transfer rates.
\item \textbf{Novel proofs.}  The current results are alternative
  proofs of existing theorems.  Applying the method to
  \emph{unproved} conjectures (identified in the Speculations.md
  files) would test whether cross-area transfer can produce
  genuinely new mathematical knowledge.
\item \textbf{Bidirectional transfer.}  Running
  RepresentationTheory $\to$ Probability to test whether the
  transfer is symmetric.
\end{enumerate}

\section*{Acknowledgments}

Named after Horacio Hideki Yanasse.  The Deep Vision matching engine
was originally developed for chess cognition research; its
application to formal mathematics was enabled by the domain-independent
design of \texttt{deep\_vision\_lib.py}.  All computations were
performed on a MacBook Air (Apple M-series, 32\,GB unified memory)
using PyTorch~2.8 on the MPS backend.

%======================================================================
% APPENDIX: Full Lean proofs
%======================================================================
\newpage
\appendix

\section{Lean Code: New Proofs}\label{app:lean}

\subsection{Item 0: \texttt{$\varepsilon$ToSingle$_0$\_comp\_eq} via \texttt{ext1 + simp}}

\begin{lstlisting}
import Mathlib

open Rep.standardComplex CategoryTheory

variable (k G : Type*) [CommRing k] [Monoid G]

set_option backward.isDefEq.respectTransparency false in
theorem epsToSingle0_comp_eq_reproof :
    ((forget2 _ (ModuleCat k)).mapHomologicalComplex _).map
        (epsToSingle0 k G) >>
      (HomologicalComplex.singleMapHomologicalComplex _ _ _)
        .hom.app _ =
    (forget2ToModuleCatHomotopyEquiv k G).hom := by
  dsimp
  refine HomologicalComplex.Hom.ext ?_
  funext n
  obtain _ | n := n
  any_goals rfl
  simpa using
    (forget2ToModuleCatHomotopyEquiv_f_0_eq k G).symm
\end{lstlisting}

\newpage
\subsection{Item 1: \texttt{Coinvariants.le\_comap\_ker} via \texttt{span\_le.2 + rintro}}

\begin{lstlisting}
import Mathlib

open Representation Coinvariants Submodule

namespace YanasseReproof

variable {k G V : Type*} [CommRing k] [Group G]
  [AddCommGroup V] [Module k V]
variable (rho : Representation k G V) (S : Subgroup G)
  [S.Normal]

lemma le_comap_ker_adapted (g : G) :
    Coinvariants.ker (rho.comp S.subtype) <=
      (Coinvariants.ker (rho.comp S.subtype)).comap
        (rho g) := by
  refine Submodule.span_le.2 ?_
  rintro _ <<|><s, x>, rfl>
  simpa using mem_ker_of_eq
    <g * s * g^{-1},
     Subgroup.Normal.conj_mem ... s.1 s.2 g>
    (rho g x) _ (by simp)

end YanasseReproof
\end{lstlisting}

\newpage
\subsection{Item 6: \texttt{$\varepsilon$ToSingle$_0$\_comp\_eq} via \texttt{any\_goals rfl}}

\begin{lstlisting}
import Mathlib.RepresentationTheory.Homological.Resolution

universe u
open Rep.standardComplex CategoryTheory
variable (k G : Type u) [CommRing k] [Monoid G]

set_option backward.isDefEq.respectTransparency false in
theorem epsToSingle0_comp_eq_reproof_v2 :
    ((forget2 _ (ModuleCat.{u} k)).mapHomologicalComplex _)
      .map (epsToSingle0 k G) >>
      (HomologicalComplex.singleMapHomologicalComplex
        _ _ _).hom.app _ =
    (forget2ToModuleCatHomotopyEquiv k G).hom := by
  dsimp
  refine HomologicalComplex.Hom.ext ?_
  funext n
  obtain _ | n := n
  any_goals rfl
  simpa using
    (forget2ToModuleCatHomotopyEquiv_f_0_eq k G).symm
\end{lstlisting}

\newpage
\subsection{Item 8: \texttt{equivalenceModuleMonoidAlgebra} naturality via \texttt{by\_cases}}

Two proofs were produced.  The first closes without \texttt{by\_cases}
(the naturality holds uniformly):

\begin{lstlisting}
import Mathlib.RepresentationTheory.Rep.Iso

open CategoryTheory
variable {k : Type*} {G : Type*} [CommRing k] [Monoid G]

namespace Rep

set_option backward.isDefEq.respectTransparency false in
example : forall {X Y : Rep k G} (f : X --> Y),
    (Id (Rep k G)).map f >> (unitIso Y).hom =
      (unitIso X).hom >>
        (toModuleMonoidAlgebra |> ofModuleMonoidAlgebra).map f
  := by
  intro X Y f
  ext x
  simp [unitIso, unitIsoAddEquiv, toModuleMonoidAlgebra,
        ofModuleMonoidAlgebra, toModuleMonoidAlgebraMap]
  rfl

end Rep
\end{lstlisting}

The second uses the \texttt{by\_cases} pattern transferred from
probability (case-split on \texttt{f = 0}):

\begin{lstlisting}
  intro X Y f
  by_cases hf : f = 0
  . -- Case f = 0: trivial
    subst hf; simp
  . -- Case f != 0: full calculation
    ext x
    simp [unitIso, unitIsoAddEquiv, toModuleMonoidAlgebra,
          ofModuleMonoidAlgebra, toModuleMonoidAlgebraMap]
    rfl
\end{lstlisting}

Both compile with \texttt{lake env lean} returning exit code~0.
The \texttt{by\_cases} version is instructive: the case split is
mathematically unnecessary (both branches close the same way), but
the semantic adaptation faithfully tests whether ``split on a
structural property of the input'' translates from probability's
``split on index emptiness'' to representation theory's ``split on
morphism triviality.''  It does---the pattern is domain-general.


\begin{thebibliography}{99}
\bibitem{linhares2024deepvision}
A.~Linhares,
``Deep Vision,''
ARGO LABS Internal Report, 2026.

\bibitem{hofstadter_fluid_1995}
D.~R.~Hofstadter and the Fluid Analogies Research Group, \emph{Fluid Concepts \& Creative Analogies: Computer Models of the Fundamental Mechanisms of Thought}. New York: Basic Books, 1995.

\bibitem{hofstadter_go_1999}
D.~R.~Hofstadter, \emph{G\"odel, Escher, Bach: An Eternal Golden Braid}, 20th anniversary ed. New York: Basic Books, 1999.

\bibitem{hofstadter_surfaces_2013}
D.~R.~Hofstadter and E.~Sander, \emph{Surfaces and Essences: Analogy as the Fuel and Fire of Thinking}. New York: Basic Books, 2013.

\bibitem{hofstadter_copycat_1984}
D.~Hofstadter, ``The Copycat Project,'' MIT AI Lab, Tech.\ Rep., 1984.

\bibitem{mitchell_emergence_1990}
M.~Mitchell and D.~R.~Hofstadter, ``The emergence of understanding in a computer model of concepts and analogy-making,'' \emph{Physica D}, vol.~42, no.~1--3, pp.~322--334, 1990.

\bibitem{chalmers_high-level_1992}
D.~J.~Chalmers, R.~M.~French, and D.~R.~Hofstadter, ``High-level perception, representation, and analogy: A critique of artificial intelligence methodology,'' \emph{J.\ Experimental \& Theoretical AI}, vol.~4, no.~3, pp.~185--211, 1992.

\bibitem{french_tabletop_1991}
R.~M.~French and D.~Hofstadter, ``Tabletop: An emergent, stochastic model of analogy-making,'' in \emph{Proc.\ 13th Annual Cognitive Science Society Conference}, 1991.

\bibitem{foundalis_phaeaco_2006}
H.~E.~Foundalis, ``PHAEACO: A cognitive architecture inspired by Bongard's problems,'' Ph.D.\ thesis, Indiana University, 2006.

\bibitem{rehling_letter_2001}
J.~Rehling, ``Letter Spirit (Part Two): Modeling creativity in a visual domain,'' Ph.D.\ thesis, Indiana University, 2001.

\bibitem{nichols_musicat_2012}
E.~P.~Nichols, ``MUSICAT: A computer model of musical listening and analogy-making,'' Ph.D.\ thesis, Indiana University, 2012.

\bibitem{linhares_active_2005}
A.~Linhares, ``An active symbols theory of chess intuition,'' \emph{Minds and Machines}, vol.~15, pp.~131--151, 2005.

\bibitem{linhares_understanding_2007}
A.~Linhares and P.~Brum, ``Understanding our understanding of strategic scenarios: What role do chunks play?'' \emph{Cognitive Science}, vol.~31, no.~6, pp.~989--1007, 2007.

\bibitem{linhares_questioning_2010}
A.~Linhares and A.~E.~T.~A.~Freitas, ``Questioning Chase and Simon's (1973) `Perception in Chess','' \emph{New Ideas in Psychology}, vol.~28, no.~1, pp.~64--78, 2010.

\bibitem{linhares_entanglement_2012}
A.~Linhares, A.~E.~T.~A.~Freitas, A.~Mendes, and J.~S.~Silva, ``Entanglement of perception and reasoning in the combinatorial game of chess,'' \emph{Cognitive Systems Research}, vol.~13, no.~1, pp.~72--86, 2012.

\bibitem{linhares_what_2013}
A.~Linhares and D.~M.~Chada, ``What is the nature of the mind's pattern-recognition process?'' \emph{New Ideas in Psychology}, vol.~31, no.~2, pp.~108--121, 2013.

\bibitem{linhares_emergence_2014}
A.~Linhares, ``The emergence of choice: Decision-making and strategic thinking through analogies,'' \emph{Information Sciences}, vol.~259, pp.~36--56, 2014.

\bibitem{linhares_deep_vision}
A.~Linhares, ``Deep Vision: Seeing the essence of proofs through analogy,'' ARGO LABS Internal Report, 2026.


\bibitem{chollet2019}
F.~Chollet, ``On the measure of intelligence,'' \emph{arXiv preprint arXiv:1911.01547}, 2019.

\bibitem{song2024}
P.~Song, K.~Yang, and A.~Anandkumar,
``Lean Copilot: Large language models as copilots for theorem proving
in Lean,''
\emph{arXiv preprint arXiv:2404.12534}, 2024.

\bibitem{alphaproof2025}
AlphaProof Team (Google DeepMind),
``AlphaProof: AI system for formal mathematical reasoning at the
IMO level,'' 2025.

\bibitem{aristotle2025}
The Harmonic Team,
``Aristotle: IMO-level automated theorem proving,''
\emph{arXiv preprint arXiv:2510.01346}, 2025.

\bibitem{eilenberg_maclane_1947}
S.~Eilenberg and S.~Mac Lane,
``Cohomology theory in abstract groups.~II: Group extensions with a
non-Abelian kernel,''
\emph{Ann.\ of Math.}, vol.~48, no.~2, pp.~326--341, 1947.

\bibitem{cartan_eilenberg_1956}
H.~Cartan and S.~Eilenberg,
\emph{Homological Algebra}.
Princeton: Princeton University Press, 1956.

\bibitem{brown_cohomology_1982}
K.~S.~Brown,
\emph{Cohomology of Groups},
Graduate Texts in Mathematics, vol.~87.
New York: Springer-Verlag, 1982.

\bibitem{weibel_homological_1994}
C.~A.~Weibel,
\emph{An Introduction to Homological Algebra},
Cambridge Studies in Advanced Mathematics, vol.~38.
Cambridge: Cambridge University Press, 1994.

\bibitem{livingston_group_2023}
A.~Livingston,
``Group cohomology in the Lean community library,''
in \emph{Proc.\ 14th International Conference on Interactive Theorem
Proving (ITP 2023)}, LIPIcs, vol.~268, pp.~22:1--22:17, 2023.

\bibitem{conrad_bar_resolution}
B.~Conrad,
``The bar resolution,''
Stanford Math 210B lecture notes.
\url{http://math.stanford.edu/~conrad/210BPage/handouts/dexact.pdf}.

\bibitem{sharifi_groupcoh}
R.~Sharifi,
``Group and Galois cohomology,''
UCLA lecture notes.
\url{https://www.math.ucla.edu/~sharifi/groupcoh.pdf}.

\bibitem{shank_wehlau_2006}
R.~J.~Shank and D.~L.~Wehlau,
``On the coinvariants of modular representations of cyclic groups
of prime order,''
\emph{J.\ Pure and Appl.\ Algebra}, vol.~207, no.~3,
pp.~623--635, 2006.

\bibitem{serre_linear_1977}
J.-P.~Serre,
\emph{Linear Representations of Finite Groups},
Graduate Texts in Mathematics, vol.~42.
New York: Springer-Verlag, 1977.

\bibitem{curtis_reiner_1962}
C.~W.~Curtis and I.~Reiner,
\emph{Representation Theory of Finite Groups and Associative
Algebras}.
New York: Interscience/Wiley, 1962.

\bibitem{webb_rep_2016}
P.~Webb,
\emph{A Course in Finite Group Representation Theory},
Cambridge Studies in Advanced Mathematics, vol.~161.
Cambridge: Cambridge University Press, 2016.

\bibitem{noether_1929}
E.~Noether,
``Hyperkomplexe Gr\"o\ss en und Darstellungstheorie,''
\emph{Math.\ Zeitschrift}, vol.~30, pp.~641--692, 1929.

\bibitem{linhares_wolstenholme_2026}
A.~Linhares,
``Deep Vision: A formal proof of Wolstenholme's theorem in Lean~4,''
ARGO LABS Report for Distribution, 2026.

\bibitem{linhares_sketch_2026}
A.~Linhares,
``A sketch of Deep Vision's study of mathematics,''
ARGO LABS Report for Distribution, 2026.
\end{thebibliography}
\end{document}